%% file: icml_watershed.tex
\documentclass[10pt,twocolumn,letterpaper]{article}
\usepackage[export]{adjustbox}
\usepackage{iccv}
\usepackage{times}
\usepackage{epsfig}
\usepackage{graphicx}
\usepackage{amsmath}
\usepackage{amssymb}
\usepackage{subfigure}

\usepackage{algorithm}
\usepackage{algorithmic}

\usepackage{blindtext}
\usepackage{color}
\usepackage{amssymb}
\usepackage{amsmath}
\usepackage[extra]{tipa}
\usepackage{soul}
\usepackage{bbm}
\usepackage{epstopdf}
\usepackage{tabularx}
\usepackage{import}
\usepackage{titling}

\DeclareMathOperator*{\argmin}{arg\,min}

\newcommand{\mapletter}{}
\newcommand{\maparrow}{\rightarrow}

\let\OLDthebibliography\thebibliography
\renewcommand\thebibliography[1]{
  \OLDthebibliography{#1}
  \setlength{\parskip}{0pt plus 0.3pt}
  \setlength{\itemsep}{0pt}
}

\usepackage[pagebackref=true,breaklinks=true,colorlinks,bookmarks=false]{hyperref}

\iccvfinalcopy 

\ificcvfinal\fi
\begin{document}

\setlength{\droptitle}{-1.7em}   
\title{Learned Watershed: End-to-End Learning of Seeded Segmentation}

\author{Steffen Wolf\thanks{Authors contributed equally}, \hspace{0.1cm} Lukas Schott\footnotemark[1], \hspace{0.1cm} Ullrich K\"othe\hspace{0.1cm} and\hspace{0.05cm} Fred Hamprecht
}

\date{HCI/IWR, Heidelberg University, 69120 Germany \\
{\tt\small firstname.lastname@iwr.uni-heidelberg.de}\\
}

\maketitle
\thispagestyle{empty}

\begin{abstract}
\noindent Learned boundary maps are known to outperform hand-crafted ones as a basis for the watershed algorithm. We show, for the first time, how to train watershed computation jointly with boundary map prediction. The estimator for the merging priorities is cast as a neural network that is convolutional (over space) and recurrent (over iterations). The latter allows learning of complex shape priors. The method gives the best known seeded segmentation results on the CREMI segmentation challenge. 
\end{abstract}

\section{Introduction}
\label{seq:Introduction}

\noindent The watershed algorithm is an important computational primitive in low-level computer vision. Since it does not penalize segment boundary length, it exhibits no shrinkage bias like multi-terminal cuts or (conditional) random fields and is especially suited to segment objects with high surface-to-volume ratio, e.g. neurons in biological images.

In its classic form, the watershed algorithm comprises three basic steps: altitude computation, seed definition, and region assignment. These steps are designed manually for each application of interest. In a typical setup, the altitude is the output of an edge detector (e.g. the gradient magnitude or the gPb detector \cite{arbelaez_11_gpb}), the seeds are located at the local minima of the altitude image, and pixels are assigned to seeds according to the drop-of-water principle \cite{watershedcuts}.

In light of the very successful trend towards learning-based image analysis, it is desirable to eliminate hand-crafted heuristics from the watershed algorithm as well. Existing work shows that learned edge detectors significantly improve segmentation quality, especially when convolutional neural networks (CNNs) are used \cite{ciresan_12_deep-em-segmentation,ronneberger_15_u-net,xie2015holistically,bai2016deep_watershed}. We take this idea one step further and propose to learn altitude estimation and region assignment {\em jointly}, in an end-to-end fashion: Our approach no longer employs an auxiliary objective (e.g. accurate boundary strength prediction), but trains the altitude function together with the subsequent region assignment decisions so that the final segmentation error is minimized directly. The resulting training algorithm is closely related to reinforcement learning.

\begin{figure}[t]
\begin{center}
   \includegraphics[width=0.95\linewidth]{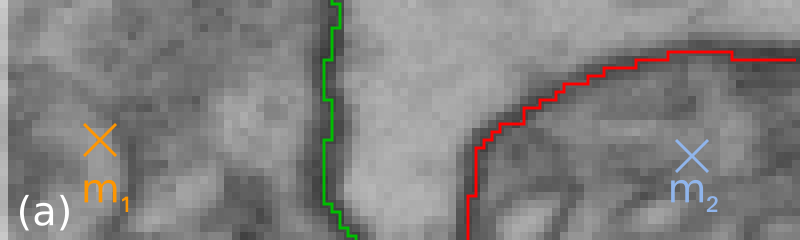}\\[0.15cm]
   \includegraphics[width=0.95\linewidth]{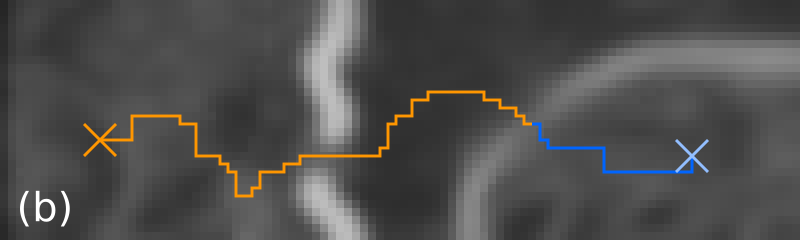}\\[0.15cm]

   \resizebox{0.8\linewidth}{2.3cm}{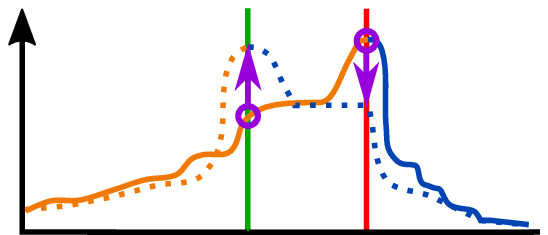}
\end{center}
   \caption{
Illustration of watershed learning. (a) Raw data with two seeds $m_1, m_2$, ground truth boundary (green), and false boundary (red) at an early training stage. (b) A hole in the altitude map causes the left region to bleed out along the orange path, meeting the blue path at a false watershed. (c) Profile of topographic distances $T$ along the two paths. Training increases the altitude at the missed edge (green) and decreases it at the false edge (red). Upon convergence, dotted paths meet at the correct location.
   }
\label{fig:fig1}
       \vspace{-0.2cm}
\end{figure}

Our method keeps the basic structure of the watershed algorithm intact: Starting from given seeds\footnote{Incorporating seed definition into end-to-end learning is a future goal of our research, but beyond the scope of this paper.}, we maintain a priority queue storing the topographic distance of candidate pixels to their nearest seed. Each iteration assigns the currently best candidate to ``its'' region and updates the queue. The topographic distance is induced by an altitude function estimated with a CNN. Crucially, and deviating from prior work, we compute altitudes {\em on demand}, allowing their conditioning on prior decisions, i.e. partial segmentations. The CNN thus gets the opportunity to learn priors for likely region shapes in the present data. We show how these models can be trained end-to-end from given ground truth segmentations using {\em structured learning}. Our experiments show that the resulting segmentations are better than those from hand-crafted algorithms or unstructured learning.

\section{Related Work}

\noindent Various authors demonstrated that learned boundary probabilities (or, more generally, boundary strengths) are superior to designed ones. In the most common setting, these probabilities are defined on the pixel grid, i.e. on the nodes of a grid graph, and serve as input of a {\em node-based} watershed algorithm. Training minimizes a suitable loss (e.g. squared or cross-entropy loss) between the predicted probabilities and manually generated ground truth boundary maps in an {\em unstructured} manner, i.e. over all pixels independently. This approach works especially well with powerful models like CNNs. In the important application of connectomis (see section \ref{sec:cremi}), this was first demonstrated by \cite{jain2007supervised}. A much deeper network \cite{ciresan_12_deep-em-segmentation} was the winning entry of the ISBI 2012 Neuro-Segmentaion Challenge \cite{isbi2012challenge}. Results could be improved further by progress in CNN architectures and more sophisticated data augmentation, using e.g. U-Nets \cite{ronneberger_15_u-net}, FusionNets \cite{quan2016fusionnet} or networks based on inception modules \cite{beier2017multicut}. Clustering of the resulting watershed superpixels by means of the GALA algorithm \cite{GALA,knowles2016rhoananet} (using altitudes from \cite{isbi2012challenge} resp. \cite{ronneberger_15_u-net}) or the lifted multicut \cite{beier2017multicut} (using altitudes from their own CNN) lead to additional performance gains.

When ground truth is provided in terms of region labels rather than boundary maps, a suitable boundary map must be created first. Simple morphological operations were found sufficient in \cite{ronneberger_15_u-net}, while \cite{beier2017multicut} preferred smooth probabilities derived from a distance transform starting at the true boundaries. Outside connectomics, \cite{bai2016deep_watershed} achieved superior results by defining the ground truth altitude map in terms of the {\em vector} distance transform, which allows optimizing the prediction's gradient direction and height separately.

Alternatively, one can employ the {\em edge-based} watershed algorithm and learn boundary probabilities for the grid graph's edges. The corresponding ground truth simply indicates if the end points of each edge are supposed to be in different segments or not. From a theoretical perspective, the distinction between node- and edge-based watersheds is not very significant because both can be transformed into each other \cite{meyer2014watersheds}. However, the algorithmic details differ considerably. Edge-based altitude learning was first proposed in \cite{fowlkes_03_learning-affinity}, who used hand-crafted features and logistic regression. Subsequently, \cite{turaga2010convolutional} employed a CNN to learn features and boundary probabilities simultaneously. Watershed superpixel generation and clustering on the basis of these altitudes was investigated in \cite{zlateski2015image}.

Learning with unstructured loss functions has the disadvantage that an error at a single point (node or edge) has little effect on the loss, but may lead to large segmentation errors: A single missed boundary pixel can cause a big false merger. Learning with {\em structured} loss functions, as advocated in this paper, avoids this by considering the boundaries in each image jointly, so that the loss can be defined in terms of segmentation accuracy rather than pointwise differences. Holistically-nested edge detection \cite{xie2015holistically,kokkinos2015pushing} achieves a weak form of this by coupling the loss at multiple resolutions using deep supervision. Such a network was successfully used as a basis for watershed segmentation in \cite{cai2016pancreas}. 
The MALIS algorithm \cite{MALIS} computes shortest paths between pairs of nodes and applies a correction to the highest edge along paths affected by false splits or mergers. This is similar to our training, but we apply corrections to root error edges as defined below. Learned, sparse reconstruction methods such as MaskExtend \cite{meirovitch2016multi} and Flood-filling networks \cite{floodfill} predict region membership for all nodes in a patch jointly, performing region growing for one seed at a time in a one-against-the-rest fashion. In contrast, our algorithm grows all seeds simultaneously and competitively.

\section{Mathematical Framework}

\noindent The watershed algorithm is especially suitable when regions are primarily defined by their boundaries, not by appearance differences. This is often the case when the goal is {\em instance} segmentation (one neuron vs. its neighbors) as opposed to semantic segmentation (neurons vs. blood vessels). In graphical model terms, pairwise potentials between adjacent nodes are crucial in this situation, whereas unary potentials are of lesser importance or missing altogether. Many real-world applications have these characteristics, see \cite{powerws} and section \ref{sec:Experiments} for examples.

We consider 4-connected grid graphs $G = (V, E)$. The input image $I: V \maparrow \mathbb{R}^D$ maps all nodes to D-dimensional vectors of raw data. A segmentation is defined by a label image $\mapletter{S}: V \maparrow \{1,2,\dots, n\}$ specifying the region index or label of each node. The ground truth segmentation is called $S^*$. Pairwise potentials (i.e. edge weights) are defined by an altitude function over the graph's edges
\begin{equation}
  f: E \maparrow \mathbb{R}
\end{equation} where higher values indicate stronger boundary evidence. Since this paper focuses on how to learn $f$, we assume that a set of seed nodes $M=\{m_1,\dots,m_n\}\subset V$ is provided by a suitable oracle (see section \ref{sec:Experiments} for details). The watershed algorithm determines $S$ by finding a mapping $\sigma:V \maparrow M$ that assigns each node to the best seed so that
\begin{equation}
\sigma(w)=m_i \quad\Rightarrow\quad S(w)=i
\end{equation}
Initially, node assignments are unknown (designated by $\lambda$) except at the seeds, where they are assumed to be correct:
\begin{equation}
\sigma_0(w) = \begin{cases}
m_i & \text{if } w=m_i\text{ with }S^*(m_i)=i \\
\lambda & \text{otherwise}
\end{cases}
\end{equation}
In this paper, we build upon the {\em edge-based} variant of the watershed algorithm \cite{meyer1994minimum, watershedcuts}. This variant is also known as {\em watershed cuts} because segment boundaries are defined by cuts in the graph, i.e. by the set of edges whose incident nodes have different labels. We denote the cuts in our solution as $\partial S$ and in the ground truth as $\partial S^*$.

Let $\Phi(m,w)$ denote the set of all paths from seed $m$ to node $w$. Then the {\em max-arc topographic distance} between $m$ and $w$ is defined as \cite{falcao2004image}
\begin{equation}
T(m,w) = \min_{\phi \in \Phi(m,w)}\, \max_{e \in \phi} f(e)
\end{equation}
In words, the highest edge in a path $\phi$ determines the path's altitude, and the path of lowest altitude determines the topographic distance. The watershed algorithm assigns each node to the topographically closest seed \cite{roerdink2000watershed}:
\begin{equation}\label{eq:closest-seed}
  \sigma(w) = \argmin_{m\in M} \,\,T(m,w)
\end{equation}
The minimum distance path from seed $m$ to node $w$ shall be denoted by $\phi_m(w)$. This path is not necessarily unique, but ties are rare and can be broken arbitrarily when $f(e)$ is a real-valued function of noisy input data.

It was shown in \cite{watershedcuts} that the resulting partitioning is equivalent to the minimum spanning forest (MSF) over seeds $M$ and edge weights $f(e)$. Thus, we can compute the watershed segmentation incrementally using Prim's algorithm: Starting from initial seeds $\sigma_0$, each iteration $k$ finds the lowest edge whose start point $u_k$ is already assigned, but whose end point $v_k$ is not
\begin{equation}
u_k,v_k =\mkern-6mu \argmin_{\substack{(u,v)\in E \\ \sigma_{k-1}(u)\ne\lambda,\,\,\sigma_{k-1}(v) = \lambda}} \mkern-6mu f(e=(u,v))\label{eq:prio_que_argmin}
\end{equation}
and propagates the seed assignment from $u_k$ to $v_k$:
\begin{equation}
\sigma_k(w) = \begin{cases}
    \sigma_{k-1}(u_k) & \text{ if } w = v_k \\
    \sigma_{k-1}(w) & \text{ otherwise }
\end{cases}\label{eq:rec_prop}
\end{equation}
In a traditional watershed implementation, the altitude $f(e)$ is a fixed, hand-designed function of the input data
\begin{equation}
f(e)=f_\text{fixed}(e|I)
\end{equation}
for example, the image's Gaussian gradient magnitude or the ``global Probability of boundary'' (gPb) detector \cite{arbelaez_11_gpb}.

\section{Joint Structured Learning of Altitude and Region Assignment}\label{sec:joint_loss}

\noindent We propose to use structured learning to train an altitude regressor $f(e)$ {\em jointly} with the region assignment procedure defined by Prim's algorithm. We will discuss two types of learnable altitude functions: $f_\text{static}$ comprises models that, once trained, only depend on the input image $I$, whereas $f_\text{dyn}$ additionally incorporates dynamically changing information about the current state of Prim's algorithm.

\subsection{Static Altitude Prediction}\label{sec:stat_alt_prediction}

\noindent To find optimal parameters $\theta$ of a model $f_\text{static}(e|I;\theta)$, consider how Prim's algorithm proceeds: It builds a MSF which assigns each node $w$ to the closest seed $\hat m=\sigma(w)$ by identifying the shortest path $\phi_{\hat m}(w)$ from $\hat m$ to $w$. Such a path can be wrong in two ways: it may cross $\partial S^*$ and thus miss a ground truth cut edge, or it may end at a false cut edge, placing $\partial S$ in the interior of a ground truth region~(see figure \ref{fig:fig1}). More formally, we have to distinguish two failure modes: (i) A node was assigned to the wrong seed, i.e. $\hat m \ne m^* = m_{S^*(w)}$ or (ii) it was assigned to the correct seed via a non-admissible path, i.e. a path taking a detour across a different region. To treat both cases uniformly, we construct the corresponding ground truth paths $\psi_{m^*}(w)$.




\begin{figure}[t]
(a)\\
\vspace{-1cm}
\begin{center}

\resizebox{0.65\linewidth}{!}{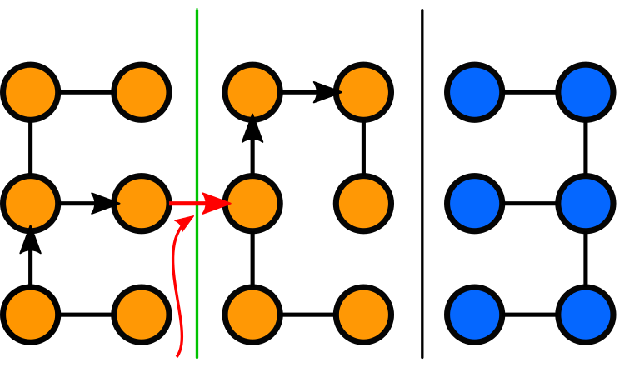}

\end{center}
(b)\\
\vspace{-1cm}
\begin{center}

\resizebox{0.65\linewidth}{!}{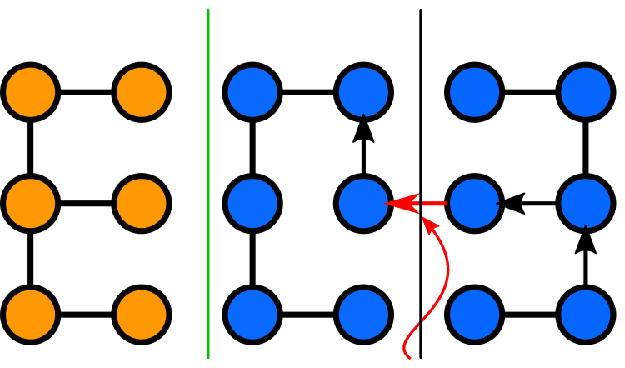}\\

\end{center}

   \caption{
   Example of root errors in the minimal spanning forest~(a) and the constrained MSF~(b) of a grid graph. Orange and blue indicate the segmentation $S$ in (a) and $S^*$ in (b). The root errors $\rho(w)$ and $\rho^*(w)$ of a wrongly labeled node $w$ are marked red, with corresponding paths $\phi_{\hat m}(w)$ and $\psi_{m^*}(w)$ depicted by arrows.
 }
    \vspace{-0.1cm}
\label{fig:false_cut}
\end{figure} %

These paths can be found by running Prim's algorithm with a modified altitude
\begin{equation}
\tilde f(e) = \begin{cases}
\infty & \text{ if }e\in\partial S^* \\
f_\text{static}(e|I; \theta) & \text{ otherwise}
\end{cases}
\end{equation}
forcing cuts in the resulting constrained MSF to coincide with $\partial S^*$~(see figure \ref{fig:false_cut}). We denote the topographic distances along $\phi_{\hat m}(w)$ and $\psi_{m^*}(w)$ as $T(\hat m, w)$ and $T^*(m^*, w)$ respectively. By construction of the MSF, $\phi$ and $\psi$ are equal for all correct nodes. Conversely, they differ for incorrect nodes, causing distance $T^*$ to exceed distance $T$. This property defines the set $V_-$ of incorrect nodes:
\begin{equation}
V_-= \{ w: T^*(m^*,w) > T(\hat m, w)\}
\end{equation}
Every incorrect path $\phi_{\hat m}(w)$ contains at least one erroneous cut edge. The first such edge shall be called the path's {\em root} error edge $\rho(w)$ and is always a missing cut. Training should increase its altitude until it becomes part of the cut set $\partial S$. The root error edge $\rho^*(w)$ of a ground truth path $\psi_{m^*}(w)$ is the first false cut edge in $\psi$ in failure mode (i) and the first edge where $\psi$ deviates from $\phi$ in mode (ii). Here, the altitude should be decreased to make the edge part of the MSF, see figure \ref{fig:false_cut}. Accordingly, we denote the sets of root edges as $E_\uparrow:=\{\rho(w):w\in V_-\}$ and $E_\downarrow:=\{\rho^*(w):w\in V_-\}$.

Since all assignment decisions in Prim's algorithm are conditioned on decisions taken earlier, the errors in any path also depend on the path's root error. Structured learning must therefore consider these errors jointly, and we argue that training updates must be derived solely from the root edges: They are the only locations whose required correction direction is unambiguously known. In contrast, we cannot even tell if subsequent errors will simply disappear once the root error has been fixed, or need updates of their own. When the latter applies, however, these edges will eventually become root errors in later training epochs, and we delay updating them to that point.

Since we need a differentiable loss to perform gradient descent, we use the perceptron loss of distance differences:
\begin{equation}
  \mathcal{L} = \sum_w T^*(m^*,w) - T(\hat m, w)
\end{equation}
Correct nodes have zero contribution since $T^*=T$ holds for them. To serve as a basis for structured learning, we transform this into a loss over altitude differences at root edges. Since topographic distances equal the highest altitude along the shortest path, we have
\begin{equation}
  T(\hat m, w) \ge f_\text{static}(\rho(w))
\end{equation}
To derive similar relations for $T^*$, consider how the constrained MSF is constructed from the unconstrained one: First, edges crossing $\partial S^*$ are removed from the MSF. Each of the resulting orphaned subgraphs is then reconnected into the constrained MSF via the lowest edge not crossing $\partial S^*$. The newly inserted edges are the root edges $\rho^*$ of all their child nodes, i.e. all nodes in the respective subgraph. Since these root edges did not belong to the original MSF, their altitude cannot be less than the maximum altitude in the corresponding child subgraph. For $w\in V_-$, it follows that
\begin{equation}
  T^*(m^*, w) = f_\text{static}(\rho^*(w))
\end{equation}
We can therefore upper-bound the perceptron loss by
\begin{equation}
  \mathcal{L}_\text{SL} = \sum_{w\in V_-} f_\text{static}(\rho^*(w)) - f_\text{static}(\rho(w)) \ge \mathcal{L}
\end{equation}
and minimize this upper bound. By rearranging the sum, the loss can be simplified into
\begin{equation}
\mathcal{L}_\text{SL}(\theta) =  \sum_{e  \in E} R(e) f_\text{static}(e|I;\theta)
 \label{eq:theoretical_loss}
\end{equation}
where we introduced a weight function counting the children of each root edge
\begin{equation}
  R(e) := \begin{cases}
        \quad \sum_{w:\, e=\rho^*(w)} 1 & \text{ if }e \in E_\downarrow\\
        -\sum_{w:\, e=\rho(w)} 1 & \text{ if }e \in E_\uparrow\\
        \quad 0     & \text{ otherwise }\label{eq:binary_reward}
    \end{cases}
\end{equation}
A training epoch of structured learning thus consists of the following steps:
\begin{enumerate}
\item Compute $f_\text{static}(e|I;\theta^{(t)})$ and $\tilde f^{(t)}(e)$ with current model parameters $\theta^{(t)}$ and determine the MSF and the constrained MSF.
\item Identify root edges and define the weights $R^{(t)}(e)$ and the loss $\mathcal{L}^{(t)}_\text{SL}(\theta)$.
\item Obtain an updated parameter vector $\theta^{(t+1)}$ via gradient descent on $\nabla_\theta \mathcal{L}^{(t)}(\theta)$ at $\theta=\theta^{(t)}$.
\end{enumerate}
These steps are iterated until convergence, and the resulting final parameter vector is denoted as $\theta_\text{SL}$.

\subsection{Relation to Reinforcement Learning}\label{sec:rel_to_RF}

\noindent In this section we compare the structured loss function $\mathcal{L}_\text{SL}$ with policy gradient reinforcement learning, which will serve as motivation for a refinement of the weighting function $R(e)$. To see the analogy, we refer to continuous control deep reinforcement learning as proposed by \cite{sutton1999policy,silver2014deterministic, lillicrap2015continuous}.

Looking at the region growing procedure from a reinforcement learning perspective, we define states as tuples $s = (e, I)$ where $e\in E$ is the edge under consideration, and the action space $\mathcal{A} := \mathbb{R}$ is the altitude to be predicted by $f_\text{static}(e|I;\theta)$. The {\em Policy Gradient Theorem} \cite{sutton1999policy} defines the appropriate update direction of the parameter vector $\theta$. In a continuous action space, it reads
\begin{equation}
    \nabla_\theta J = \nabla_\theta \sum_{s} d^\pi(s) \int_\mathcal{A} \pi(a| s; \theta)\,Q^\pi(s,a) \,\, da
\end{equation}
where $J$ is the performance to be optimized, $d^\pi(s)$ the discounted state distribution, $\pi$ the policy to be learned, and $Q$ the action-value function estimating the discounted expected future reward
\begin{align}\label{eq:q-function}
Q^\pi(s,a) =&\mathbb{E}_\pi \Big[ \sum_{t=0}^T \gamma^t r^t \Big| a_0=a, s_0=s; \pi\Big]
\end{align}
In our case, the state distribution reduces to $d^\pi(s)= \frac{1}{|V|}$ because Prim's algorithm reaches each edge exactly once. Inserting our deterministic altitude prediction
\begin{equation}
   \pi(a | s) := f_{\text{static}}(s | I; \theta) \, \, \, \delta(a - f_{\text{static}}(s | I; \theta)),
\end{equation}
where $\delta$ is the Dirac distribution, we get
\begin{equation}
    \nabla_\theta J = \frac{1}{|V|} \nabla_\theta \sum_s   f_{\text{static}}(s|I; \theta)\, Q^\pi(s,a).\label{eq:RL_loss}
\end{equation}
Comparing equation (\ref{eq:RL_loss}) with equation (\ref{eq:theoretical_loss}), we observe that $\nabla_\theta J \sim \nabla_\theta \mathcal{L}_{\text{SL}}$, where our weights $R(e)$ essentially play the role of the action-value function $Q$. This suggests to introduce a {\em discount factor} in $R(e)$. To do so, we replace the temporal differences $t$ between states in (\ref{eq:q-function}) with tree distances $\text{dist}(w, \rho(w))$ or $\text{dist}(w, \rho^*(w))$ counting the number of edges between node $w$ and its root edge. This gives the discounted weights
\begin{equation}
  R_{\text{RL}}(e) := \begin{cases}
       \sum\limits_{w:\, e=\rho^*(w)} \gamma^{\text{ dist}(w,\,\rho^*\mkern-2mu(w))} & \text{ if }e \in E_\downarrow\\
        \sum\limits_{w:\, e=\rho(w)} -\gamma^{\text{ dist}(w,\,\rho(w))} & \text{ if }e \in E_\uparrow\\
        \qquad\qquad 0     & \text{ otherwise }
    \end{cases}
    \label{eq:R_RL}
\end{equation}
with discount factor $ 0\le \gamma \le 1$ to be chosen such that $\gamma^\text{dist}$ decays roughly according to the size of the CNNs receptive field. Substituting $R_{RL}(e)$  for $R(e)$ in (\ref{eq:theoretical_loss}) significantly improves convergence in our experiments. This analogy further motivates the application of current deep reinforcement training methods as described in section \ref{sec:NN_training}.

\subsection{Dynamic Altitude Prediction}\label{sec:dyn_alt_prediction}

\noindent In every iteration, region growing according to Prim's algorithm only considers edges with exactly one end node located in the already assigned set. This offers the possibility to delay altitude estimation to the time when the values are actually needed. On-demand altitude computations can take advantage of the partial segmentations already available to derive additional {\em shape} clues that help resolving difficult assignment decisions.

\textbf{Relative Assignments:} To incorporate partial segmentations, we remove their dependence on the incidental choice of label values by means of {\em label-independent projection}. Consider an edge $e=(u,v)$ where node $u$ is assigned to seed $m$ and node $v$ is unassigned. We now construct a labeling relative to $m$, distinguishing nodes assigned to $m$ (``me'' region), to another seed (``them'') and unassigned (``nobody''). Relative labelings are represented by a standard 1-of-3 coding:
\begin{equation}
  \mathcal{P}(w\,|\,m,\sigma)=
   \begin{cases}
     (1,0,0) & \text{ if }\sigma(w) = m\,\text{ (me)} \\
     (0,1,0) & \text{ if }\sigma(w) = \lambda\,\,\,\text{ (nobody)} \\
     (0,0,1) & \text{ otherwise\quad\,\,\,\,\,(them)}
  \end{cases}
\end{equation}
In practice, we process relative labelings by adding a new branch to our neural network that receives $\mathcal{P}$ as an input, see section \ref{sec:NN_Realization} for details.

\textbf{Non-Markovian modeling:} Another potentially useful cue is afforded by the fact that Prim's algorithm
propagates the assignments recursively.
Thus during every evaluation of $f$ the complete history from previous iterations along the growth paths $\phi_m$ can be incorporated.
We encode the history $ \mathcal{H}\,:\,V\rightarrow\mathbb{R}^r$ about past assignment decisions as an $r$-dimensional vector in each node.
In practice, we incorporate history by adding a recurrent layer to our neural network.

We introduce the {\em dynamic} altitude predictions: 
\begin{align}
\begin{split}
f(e=(u, &v)), \mathcal{H}(v) = \\ & f_\text{dyn}(e\,|\ I, \mathcal{P}(.\,|\,\sigma(u),\sigma), \mathcal{H}(u); \theta_\text{dyn})
\end{split}
\end{align}
that receives the relative assignments projection $\mathcal{P}$ and $u$'s hidden state $\mathcal{H}(u)$ as an additional input and outputs both the edge's altitude $f(e)$ and $v$'s hidden state $\mathcal{H}(v)$:
This variant of the altitude estimator performs best in our experiments.
The emergent behavior of our models suggests that the algorithm uses history to adjust local diffusion parameters such as preferred direction and ``viscosity'', similar to the viscous watershed transform described in \cite{viscous2005}.

\begin{figure}[tbp]
 \begin{center}
   \includegraphics[width=0.85\linewidth]{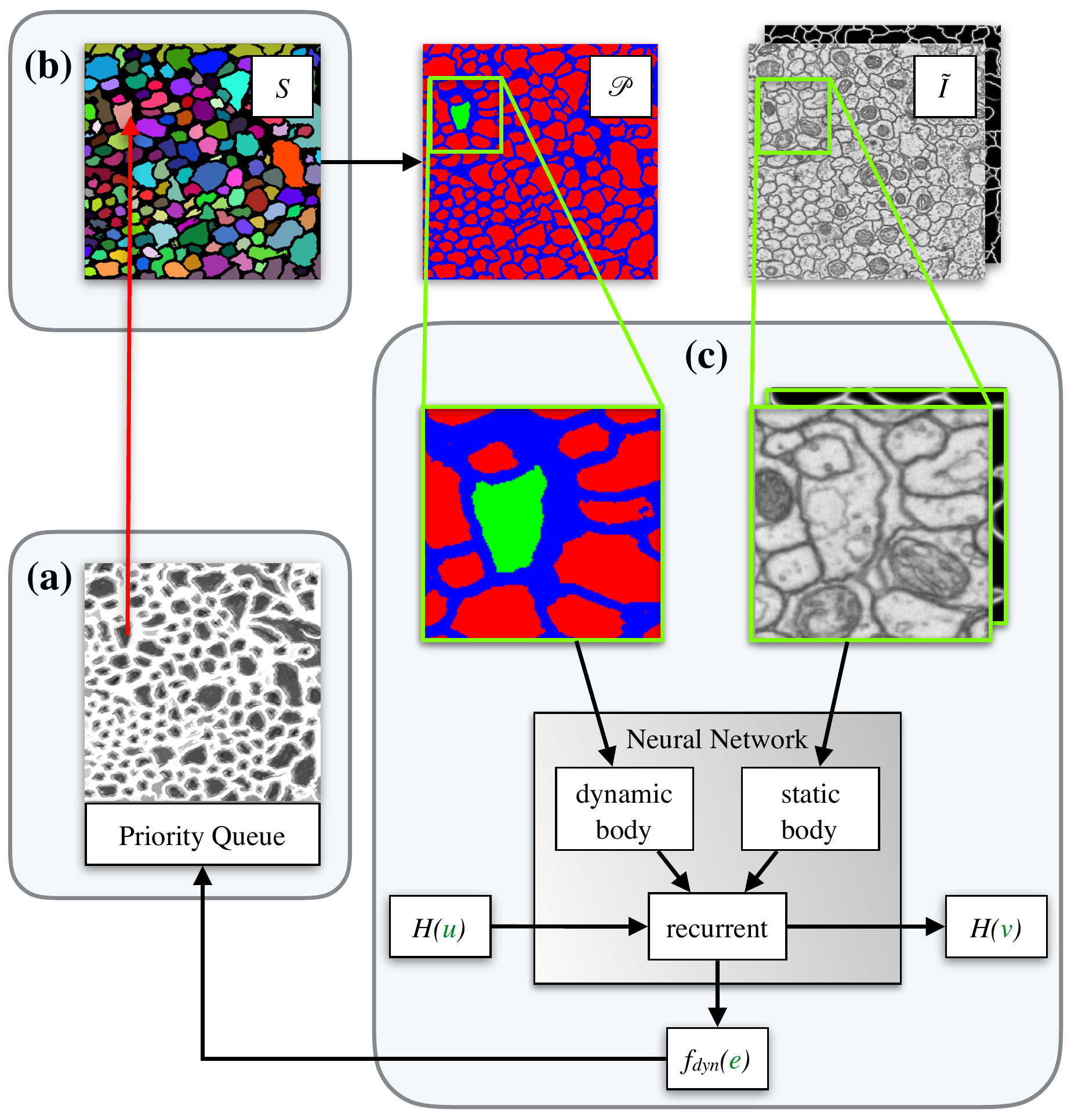}
\end{center}
   \caption{Overview implementation of learned watershed algorithm with neural network and priority queue. In each iteration the minimal edge according to equation (\ref{eq:prio_que_argmin}) is found using a priority queue (a) and the region label is propagated (b), which updates the projection $\mathcal{P}$.
    For all unassigned edges that are not in the priority queue and need to be considered by Prim's algorithm in the next iteration, the altitude $f_{\text{dyn}}(e)$ is evaluated using the dynamic edge prediction network~(c).}
    \vspace{-0.4cm}
\label{fig:overview}
\end{figure}

\section{Methods}

\subsection{Neural Network Architecture}\label{sec:NN_Realization}

\begin{figure}[tbp]
 \begin{center}
   \includegraphics[width=0.99\linewidth]{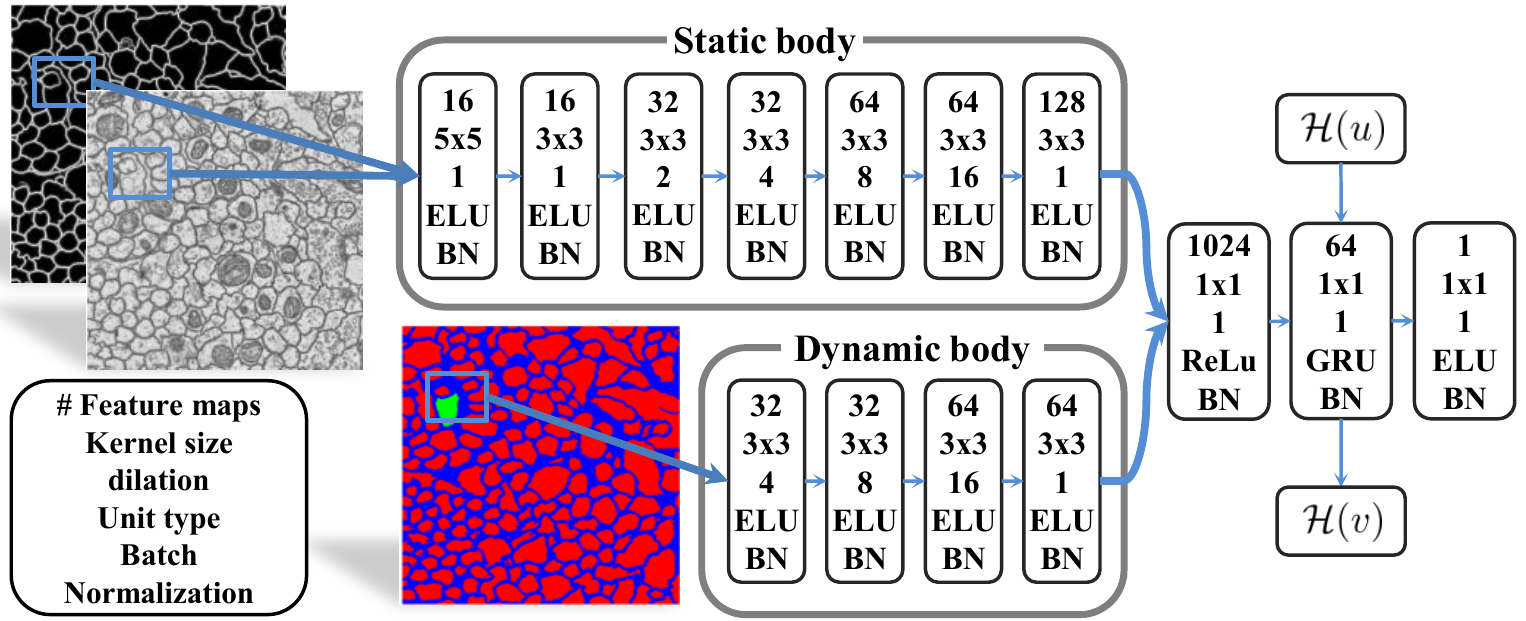}
\end{center}
   \caption{Network architecture: The static convolutional body extracts features from the raw input and edge detector output. The more shallow dynamic body processes the interactions of different region projections $\mathcal{P}$. These informations are combined in a fully connected layer and set into a temporal context using a recurrent GRU layer. The network output is the priority of the edge towards the pixel at the center of the field of view.}
   \vspace{-0.4cm}
\label{fig:NN_arch}
\end{figure}

\noindent Our network architecture builds mainly on the work of Yu and Koltun~\cite{DBLP:journals/corr/YuK15} who introduced dilated convolutions to achieve dense segmentations and systematically aggregate multi-scale contextual information without pooling operations.
We split our network into two convolutional branches~(see Figure \ref{fig:NN_arch}): The upper branch processes the static input $I$, and the lower one the dynamic input $\mathcal{P}(\cdot)$.
Since the input of the upper branch doesn't change during prediction, its network activations can be precomputed for all edges, leading to a significant speed-up.
We choose gated recurrent units (GRU) instead of long short-term memory (LSTM) in the recurrent network part, because GRUs have no internal state and get all history from the hidden state vector $\mathcal{H}(\cdot)$, saving on memory and bookkeeping.

\subsection{Training Methods}\label{sec:NN_training}

\noindent \textbf{Augmenting the Input Image:}
We noted above that structured learning is superior because it considers edges jointly. However, it can only rely on the sparse training sets $E_\downarrow\cup E_\uparrow$. In contrast, unstructured learning can make use of all edges and thus has a much bigger training set. This means that more powerful predictors, e.g. much deeper CNNs, can be trained, leading to more robust predictions and bigger receptive fields.

To combine the advantages of both approaches, we propose to augment the input image $I$ with an additional channel holding node boundary probabilities predicted by an unstructured model $g(w|I;\theta_\text{UL})$:
\begin{equation}\label{eq:augmented-image}
  \tilde{I} := [I \quad g] : V \rightarrow \mathbb{R}^{D+1}
\end{equation}
We train the CNN $g$ separately beforehand and replace $I$ with the {\em augmented input} $\tilde I$ everywhere in $f_\text{static}$ and $f_\text{dyn}$. This simplifies structured learning because the predictor only needs to learn a refinement of the already reasonable altitudes in $g$. In principle, one could even train $f$ and $g$ jointly, but the combined model is too big for reliable optimization.

\textbf{Training Schedule:}
Taking advantage of the close relationship with reinforcement learning, we adopt the asynchronous update procedure proposed by \cite{A3C}. Here, independent workers fetch the current CNN parameters $\theta$ from the master and compute loss gradients for randomly selected training images in parallel. The master then applies these updates to the parameters in an asynchronous fashion. We found experimentally, that this lead to faster and more stable convergence than sequential training.

In order to train the recurrent network part, we replace the standard temporal input ordering with the succession of edges defined by the paths $\phi$ and $\psi$. In a sense, backpropagation in time thus becomes backpropagation along the minimum spanning forest. 

\section{Experiments and Results}
\label{sec:Experiments}
\noindent Our experiments illustrate the performance of our proposed end-to-end trainable watershed in combination with static and dynamic altitude prediction. To this end, we compare with standard watershed and power watershed algorithms~\cite{powerws} on statically trained CNNs according to \cite{beier2017multicut}, see section \ref{sec:artificial_data}. Furthermore we show in section \ref{sec:cremi} that the learned watershed surpasses the state-of-the-art segmentation in an adapted version of the CREMI Neuron Segmentation Challenge \cite{cremi}.

\subsection{Experimental Setup and Evaluation Metrics}

\noindent \textbf{Seed Generation Oracle:} \label{sec:seed_oracle}%
All segmentation algorithms start at initial seeds $M$ which are here provided by a ``perfect'' oracle.
In our experiments, this oracle uses the ground truth segmentation to select one pixel with maximal $L_2$ distance to the region boundary per ground truth region.

\textbf{Segmentation Metrics:}
In accordance with the CREMI challenge \cite{cremi}, we use the following segmentation metrics:
The Rand score $V^{\text{Rand}}$ measures the probability of agreement between segmentation $S$ and ground truth $S^*$ w.r.t. a randomly chosen node pair $w, w'$. Two segmentations agree if both assign $w$ and $w'$ to the same region or to different regions. The Rand error $\text{ARAND}=1-V^{\text{Rand}}$ is the opposite, so that smaller values are better.

The {\em Variation of Information(VOI)} between $S$ and $S^*$ is defined as $VOI(S;S^*) = H(S|S^*) + H(S^*|S),$ where $H$ is the conditional entropy \cite{VOI}. To distinguish split errors from merge errors, we report the summands separately as $\text{VOI}^{\text{SPLIT}} = H(S|S^*)$ and $\text{VOI}^{\text{MERGE}} = H(S^*|S)$

\subsection{Artificial Data}\label{sec:artificial_data}

\begin{figure}
   \includegraphics[width=\linewidth]{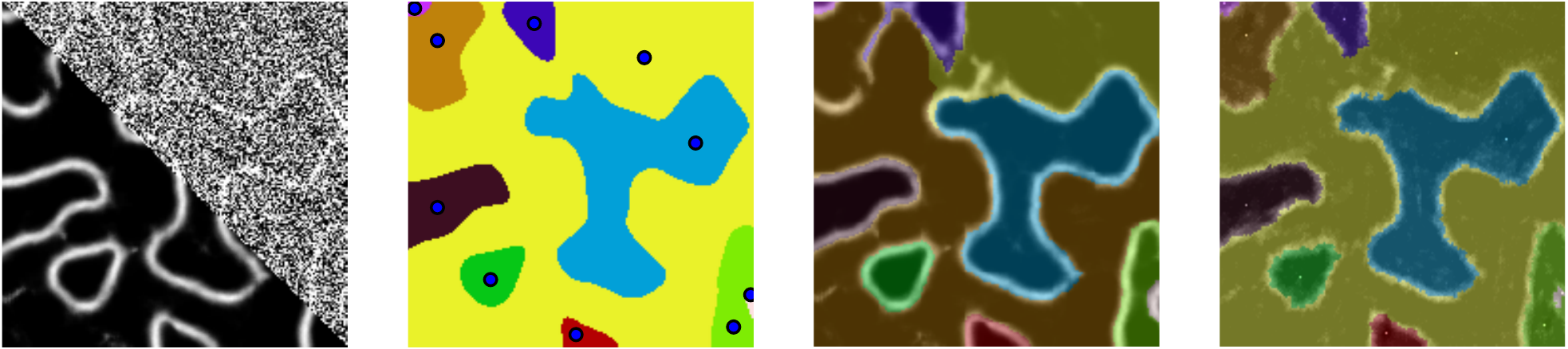}
   \caption{Artificial data example. a) Raw with $\sigma_\text{noise} = 0.6$ and prediction of baseline CNN. b) Ground truth. c) The brown region leaks out when standard watershed runs on top of baseline CNN. d) Our algorithm uses learned shape priors to close boundary gaps.}
\label{fig:toy_example0}
\vspace{-0.4cm}
\end{figure}

\noindent \textbf{Dataset:} In order to compare our models $f_\text{static}$ and $f_\text{dyn}$ with solutions based on unstructured learning, we create an artificial segmentation benchmark dataset with variable difficulty. 
First, we generate an edge image via the zero crossing of a 2D Gaussian process. This image is then smoothed with a Gaussian filter and corrupted with Gaussian noise at $\sigma_\text{noise}\in\{0.3, 0.6, 0.9\}$. For each $\sigma_{\text{noise}}$, we generate 1900 training images and 100 test images of size 252x252. One test image with corresponding ground truth and results is shown in figure~\ref{fig:toy_example0}.

\textbf{Baseline:}
We choose a recent edge detection network from \cite{DBLP:journals/corr/YuK15} to predict boundaries between different instances in combination with standard watershed (WS) and Power Watershed (PWS) \cite{powerws} to generate an instance segmentation. Since these algorithms work best on slightly smoothed inputs, we apply Gaussian smoothing to the CNN output. The optimal smoothing parameters are determined by grid search on the training dataset. Additionally, we apply all watershed methods directly to smoothed raw image and report their overall best result as \textit{RAW + WS}.

\textbf{Performance:}
The measured segmentation errors of all algorithms are shown in table \ref{tab:toy-scores}. Observed differences in performance mainly indicate how well each method handles low-contrast edges and narrow gaps between regions. The structurally trained watersheds outperform the unstructured baselines, because our loss function $\mathcal{L}_{\text{SL}}$ heavily penalizes the resulting segmentation errors. In all experiments, the {\em dynamic prediction} function $f_{\text{dyn}}$ has the best performance, due to its superior modeling power. It can identify holes and close most contours correctly because it learns to derive shape and contingency clues from monitoring intermediate results during the flooding process. A representative example of this effect is shown in figure \ref{fig:toy_example0}.

\begin{table}[h]
\begin{center}
\begin{tabular}{|l|c|c|c|r|}
\hline
        ARAND       & $\sigma_{\text{noise}} = 0.3$ & $\sigma_{\text{noise}} = 0.6$ & $\sigma_{\text{noise}} = 0.9$ \\
\hline\hline
    $\mathcal{L}_{\text{SL}}+ f_\text{dyn}$ & \textbf{5.8} $\pm$ \textbf{0.8} & \textbf{12.5} $\pm$ \textbf{1.7} & \textbf{32.2} $\pm$ \textbf{1.8} \\
    $\mathcal{L}_{\text{SL}}+ f_\text{static}$ & 6.4 $\pm$ 0.9 & 13.8 $\pm$ 1.6  & 32.4 $\pm$ 2.2 \\
    NN + WS & 6.5 $\pm$ 0.8 & 14.9 $\pm$ 3.6 & 33.4 $\pm$ 1.7 \\
    NN + PWS & 6.5 $\pm$ 0.8 & 14.9 $\pm$ 1.7 & 33.2 $\pm$ 1.7 \\
    RAW + WS & 24.0 $\pm$ 1.6 & 41.9 $\pm$ 1.8 & 55.0 $\pm$ 1.8 \\
\hline
\end{tabular}\vspace{1mm}
\caption{Quality of the segmentation results on our artificial dataset. Reported lowest error for all parameters of baseline watersheds based on the rand error and a 2 pixel boundary distance tolerance.}\label{tab:toy-scores}
\end{center}
\vspace{-0.6cm}
\end{table}

\subsection{Neurite Segmentation}\label{sec:cremi}


\noindent \textbf{Dataset:}
The MICCAI Challenge on Circuit Reconstruction from Electron Microscopy Images~\cite{cremi} contains 375 fully annotated slices of electron microscopy images $I$ (of resolution 1250x1250 pixels). Part of a data slice is displayed in figure \ref{fig:cy_full} top. Since the test ground truth segmentation has not been disclosed, we generate a new train/test split from the 3 original challenge training datasets by spltnting them into 3x75 z-continuous training- and 3x50 z-continuous test blocks.


Ideally, we would compare with \cite{beier2017multicut} whose results define the state-of-the-art on the CREMI Challenge at time of submission. However, their pipeline, as described in their supplementary material, optimizes 2D segmentations jointly across multiple slices with a complex graphical model, which is beyond the scope of this paper.

Instead, we isolate the 2D segmentation aspect by adapting the challenge in the following manner: We run each segmentation algorithm with fixed ground truth seeds (see section \ref{sec:seed_oracle}) and evaluate their results on each $z$-slice separately.
The restriction to 2D evaluation requires a slight manual correction of the ground truth: The ground truth accuracy in $z$-direction is just $\pm 1$ slice. The official 3D evaluation scores compensate for this by ignoring deviations of $\pm 1$ pixels in $z$-direction. Since this trick doesn't work in 2D, we remove 4 regions with no visual evidence in the image and all segments smaller than 5 pixels. Boundary tolerances in the x-y plane are treated as in the official CREMI scores where deviations from the true boundary are ignored if they do not exceed 6.25 pixels.

\begin{figure*}[t]
\vspace{-1cm}
\begin{center}
    \begin{tabular}{rccc}
\rotatebox{90}{\hspace{-1mm}Ground Truth}&
      \includegraphics[height=50pt]{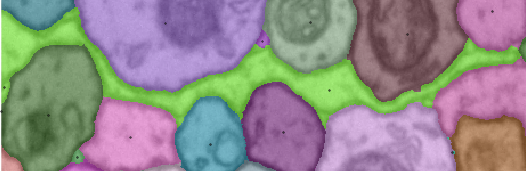} & 
    \includegraphics[height=50pt]{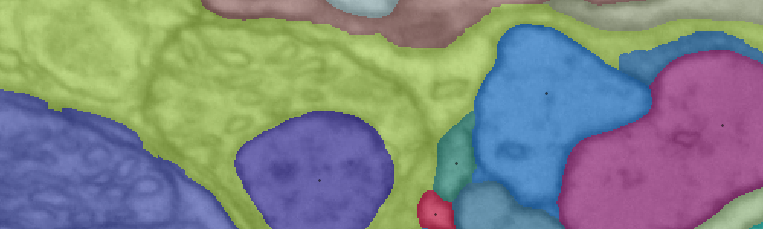} &
      \includegraphics[height=50pt]{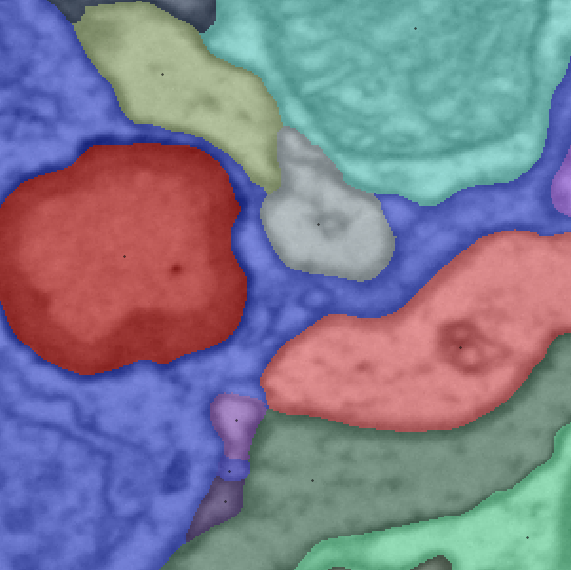}\\
      \rotatebox{90}{\hspace{3mm}DTWS}&
      \includegraphics[height=50pt]{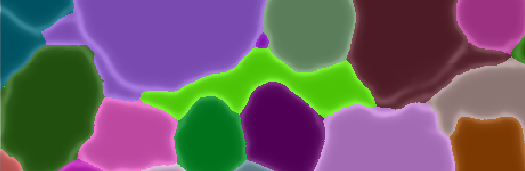} &
    \includegraphics[height=50pt]{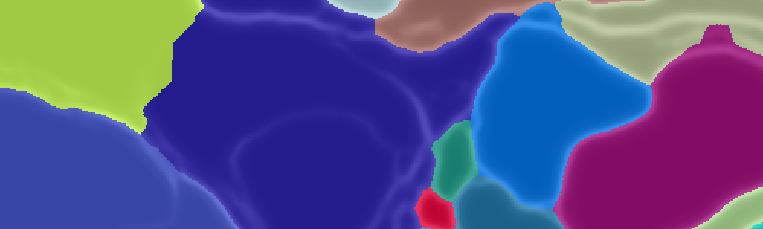}&
      \includegraphics[height=50pt]{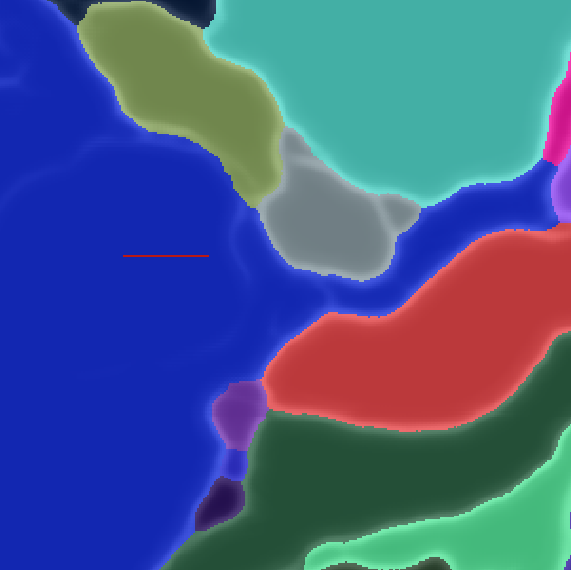}\\
      \rotatebox{90}{Learned WS}&
      \subfigure[our method follows long thin neurites]{\includegraphics[height=50pt]{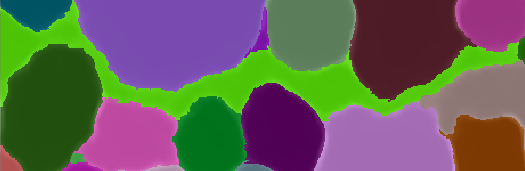}\label{fig:cy_ex_thin}}&
    \subfigure[it finds weak boundaries using shape priors]{\includegraphics[height=50pt]{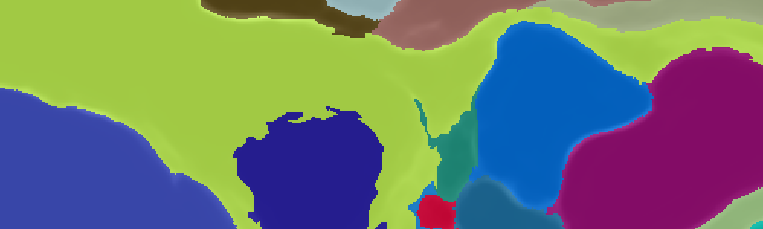} \label{fig:cy_ex_round}}&
      \subfigure[failure case]{\includegraphics[height=50pt]{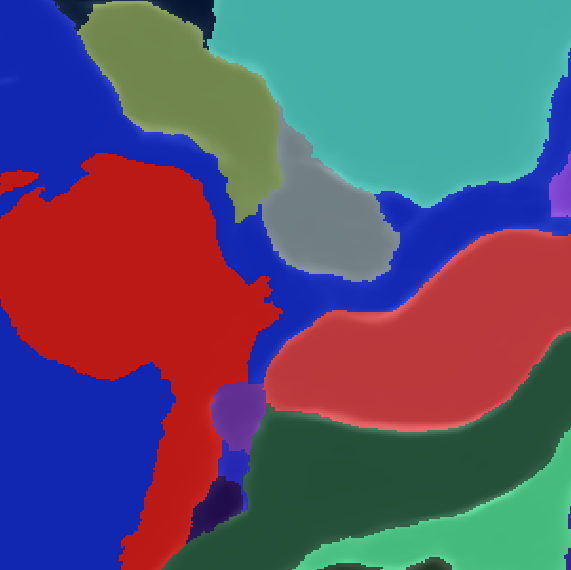}\label{fig:cy_ex_completion}}
  \end{tabular}
\end{center}
\vspace{-0.2cm}
  \caption{Detailed success and failure cases of our method.}\label{fig:cy_ex_all}
\vspace{-0.4cm}
\end{figure*}

\textbf{Baseline:}
We compare the Learned Watershed performance against the Power Watershed\cite{powerws}, Viscous Watershed \cite{viscous2005}, RandomWalker\cite{grady2006random}, Stochastic Watershed\cite{angulo2007stochastic} and Distance Transform Watershed\cite{beier2017multicut}. The boundary probability prediction $g$ (the same $g$ as in equation (\ref{eq:augmented-image})) was provided by a deep CNN trained with an unstructured loss-function. In particular, the Distance Transform Watershed (DTWS) and the prediction $g$ were used to produce the current state-of-the-art on the CREMI challenge.
To obtain the DTWS, one thresholds $g$, computes a distance transform of the background, i.e. the non-boundary pixels and runs the watershed algorithm on the inverted distances.
According to \cite{beier2017multicut}, this is the best known heuristic to close boundary gaps in these data, but requires manual parameter tuning.
We found the parameters of all baseline algorithms by grid search using the training dataset.
To ensure fair comparison, we start region growing from ground truth seeds in all cases. Our algorithm takes the augmented image $\tilde{I}$ from equation (\ref{eq:augmented-image}) as input and learns how to close boundary gaps.

\textbf{Comparison to state-of-the-art:}
We show the 2D CREMI segmentation scores in table \ref{tab:cremi_scores}. It is evident that the learned watershed transform significantly outperforms DTWS in both ARAND and VOI score. Quantitatively, we find that the flooding patterns and therefore the region shapes of the learned watershed prefer to adhere to biologically sensible structures. We illustrate this with our results on one CREMI test slice in figure \ref{fig:cy_full}, as well as specific examples in figure \ref{fig:cy_ex_all}. We find throughout the dataset that especially \textit{thin processes}, as depicted in fig. \ref{fig:cy_ex_thin} left, are a strength of our algorithm. Biologically sensible \textit{shape completions} can also be found for roundish objects and is particularly noticeable when boundary evidence is weak, as shown in Fig. \ref{fig:cy_ex_round} center. However, in rare cases, we find incorrect shape completions (see Fig. \ref{fig:cy_ex_completion} right), mainly in areas of weak boundary evidence. It stands to reason that these errors could be fixed by providing more training data.

\begin{figure}[b]
  \vspace{-0.4cm}
  \includegraphics[width=1\linewidth,trim={0 0.75cm 0 0},clip,right]{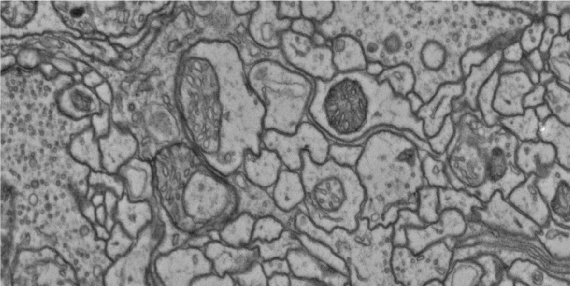}\label{fig:cremi_raw}
  \includegraphics[width=1\linewidth,trim={0 0.75cm 0 0},clip,center]{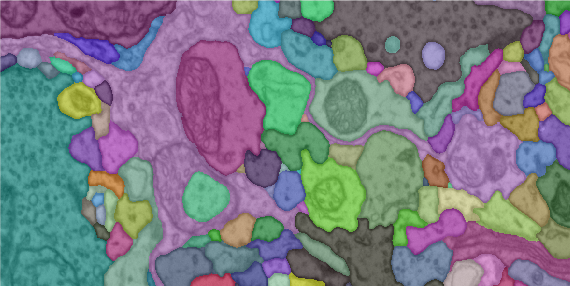}\\[0.5mm]
  \includegraphics[width=1\linewidth,trim={0 0.75cm 0 0},clip,center]{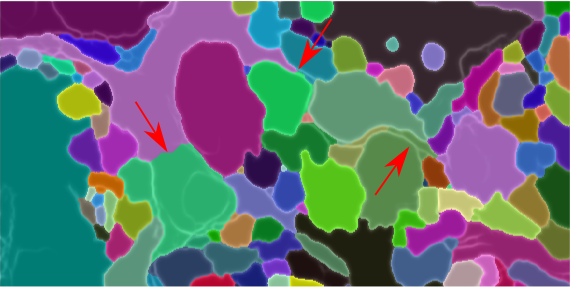}\\[0.5mm]
  \includegraphics[width=1\linewidth,trim={0 0.75cm 0 0},clip,center]{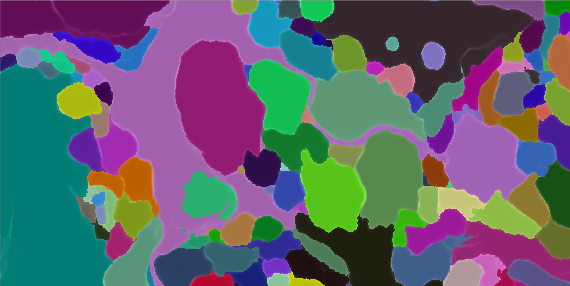}\\[0.5mm]
  \caption{From top: Raw data. Ground truth. Result of distance transform WS (arrows point out major errors). Our algorithm.}\label{fig:cy_full}
\end{figure}


\begin{table}[b]
\vspace{-0.4cm}
\centering
\footnotesize
        \begin{tabular}{|l|c|c|c|}
            \hline
                & ARAND & VOI split & VOI merge \\
            \hline
                PowerWS         & 0.122 $\pm$ 0.003     &  0.340 $\pm$ 0.031    &  0.180 $\pm$ 0.019 \\ 
                ViscousWS       & 0.093 $\pm$ 0.003     &  0.328 $\pm$ 0.030    &  0.069 $\pm$ 0.003\\ 
                RandomWalker    & 0.103  $\pm$ 0.004    &  0.355 $\pm$ 0.037    &  0.060 $\pm$ 0.004  \\
                Stochastic WS   & 0.193  $\pm$ 0.012    &  0.612 $\pm$ 0.080    &  0.077 $\pm$ 0.004 \\ 
                DTWS            & 0.085 $\pm$ 0.001     &  0.320 $\pm$ 0.029    &  0.070 $\pm$ 0.005\\
                Learned WS      & \textbf{0.082} $\pm$ \textbf{0.001}   &  \textbf{0.319} $\pm$ \textbf{0.030} & \textbf{0.057} $\pm$ \textbf{0.004} \\
            \hline
        \end{tabular}\vspace{1mm}
  \caption{CREMI segmentation metrics evaluated on 2D slices:
  The Variation of Information between a predicted segmentation and ground truth~(lower is better) and the Adapted Rand Error (lower is better) \cite{isbi2012challenge}.}\label{tab:cremi_scores}
\end{table}

\section{Conclusion}

\noindent This paper proposes an end-to-end learnable seeded watershed algorithm that performs well an artificial data and neurosegmentation EM images. We found the following aspects to be critical success factors: First, we train a very powerful CNN to control region growing. Second, the CNN is trained in a structured fashion, allowing it to optimize segmentation performance directly, instead of treating pixels independently. Third, we improve modeling power by incorporating dynamic information about the current state of the segmentation. Specifically, feeding the current partial segmentation into the CNN provides shape clues for the next assignment decision, and maintaining a latent history along assignment paths allows to adjust growing parameters locally. We demonstrate experimentally that the resulting algorithm successfully solves difficult configurations like narrow region parts and low-contrast boundaries, where previous algorithms fail. In future work, we plan to include seed generation into the end-to-end learning scheme.



\cleardoublepage

\newenvironment{changemargin}[2]{%
\begin{list}{}{%
\setlength{\topsep}{0pt}%
\setlength{\leftmargin}{#1}%
\setlength{\rightmargin}{#2}%
\setlength{\listparindent}{\parindent}%
\setlength{\itemindent}{\parindent}%
\setlength{\parsep}{\parskip}%
}%
\item[]}{\end{list}}

\setlength{\voffset}{-1cm}
\setlength{\textheight}{1.1\textheight}
{\small
\bibliographystyle{ieee}
\bibliography{watershed}
}

\end{document}

%% file: image/fig1/fig1_plot.eps_tex
\begingroup%
  \makeatletter%
  \providecommand\color[2][]{%
    \errmessage{(Inkscape) Color is used for the text in Inkscape, but the package 'color.sty' is not loaded}%
    \renewcommand\color[2][]{}%
  }%
  \providecommand\transparent[1]{%
    \errmessage{(Inkscape) Transparency is used (non-zero) for the text in Inkscape, but the package 'transparent.sty' is not loaded}%
    \renewcommand\transparent[1]{}%
  }%
  \providecommand\rotatebox[2]{#2}%
  \ifx\svgwidth\undefined%
    \setlength{\unitlength}{163.85511278bp}%
    \ifx\svgscale\undefined%
      \relax%
    \else%
      \setlength{\unitlength}{\unitlength * \real{\svgscale}}%
    \fi%
  \else%
    \setlength{\unitlength}{\svgwidth}%
  \fi%
  \global\let\svgwidth\undefined%
  \global\let\svgscale\undefined%
  \makeatother%
  \begin{picture}(1,0.39812439)%
    \put(0,0){\includegraphics[width=\unitlength]{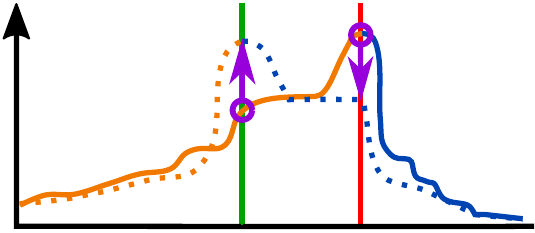}}%
    \put(0.05987469,0.34440324){\color[rgb]{0,0,0}\makebox(0,0)[lb]{\smash{T}}}%
    \put(0.12147815,0.1693449){\color[rgb]{0.94901961,0.4745098,0}\makebox(0,0)[lb]{\smash{$T(m_1, \cdot)$}}}%
    \put(0.73569273,0.17667634){\color[rgb]{0,0.26666667,0.70196078}\makebox(0,0)[lb]{\smash{$T(m_2, \cdot)$}}}%
    \put(0.76455556,0.33834842){\color[rgb]{0.59215686,0,0.85882353}\makebox(0,0)[lb]{\smash{loss gradient}}}%
    \put(0.76455556,0.26834842){\color[rgb]{0.59215686,0,0.85882353}\makebox(0,0)[lb]{\smash{$\nabla \mathcal{L}_\text{SL}$}}}%

    \put(-0.05449019,0.02278732){\color[rgb]{0,0,0}\makebox(0,0)[lb]{\smash{(c)}}}%
  \end{picture}%
\endgroup%

%% file: image/training/drawing.eps_tex
\begingroup%
  \makeatletter%
  \providecommand\color[2][]{%
    \errmessage{(Inkscape) Color is used for the text in Inkscape, but the package 'color.sty' is not loaded}%
    \renewcommand\color[2][]{}%
  }%
  \providecommand\transparent[1]{%
    \errmessage{(Inkscape) Transparency is used (non-zero) for the text in Inkscape, but the package 'transparent.sty' is not loaded}%
    \renewcommand\transparent[1]{}%
  }%
  \providecommand\rotatebox[2]{#2}%
  \ifx\svgwidth\undefined%
    \setlength{\unitlength}{177.43466034bp}%
    \ifx\svgscale\undefined%
      \relax%
    \else%
      \setlength{\unitlength}{\unitlength * \real{\svgscale}}%
    \fi%
  \else%
    \setlength{\unitlength}{\svgwidth}%
  \fi%
  \global\let\svgwidth\undefined%
  \global\let\svgscale\undefined%
  \makeatother%
  \begin{picture}(1,0.61404003)%
    \put(0,0){\includegraphics[width=\unitlength]{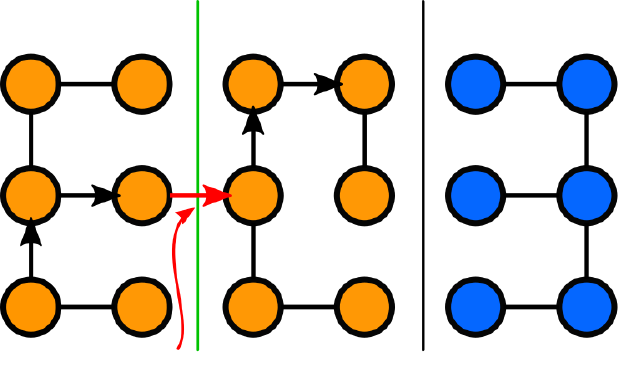}}%
    \put(0.04895871,0.0103251){\color[rgb]{1,0,0}\makebox(0,0)[lb]{\smash{root error $\rho(w)$: missing cut}}}%
    \put(0.33345982,0.56988725){\color[rgb]{0,0.76078431,0}\makebox(0,0)[lb]{\smash{$\partial S^*$}}}%
    \put(0.70220726,0.56988725){\color[rgb]{0,0,0}\makebox(0,0)[lb]{\smash{$\partial S$}}}%
    \put(0.03041218,0.10642789){\color[rgb]{0,0,0}\makebox(0,0)[lb]{\smash{$\hat{m}$}}}%
    \put(0.57244834,0.46764071){\color[rgb]{0,0,0}\makebox(0,0)[lb]{\smash{$w$}}}%
  \end{picture}%
\endgroup%

%% file: image/training/drawing_hal.eps_tex
\begingroup%
  \makeatletter%
  \providecommand\color[2][]{%
    \errmessage{(Inkscape) Color is used for the text in Inkscape, but the package 'color.sty' is not loaded}%
    \renewcommand\color[2][]{}%
  }%
  \providecommand\transparent[1]{%
    \errmessage{(Inkscape) Transparency is used (non-zero) for the text in Inkscape, but the package 'transparent.sty' is not loaded}%
    \renewcommand\transparent[1]{}%
  }%
  \providecommand\rotatebox[2]{#2}%
  \ifx\svgwidth\undefined%
    \setlength{\unitlength}{177.43466034bp}%
    \ifx\svgscale\undefined%
      \relax%
    \else%
      \setlength{\unitlength}{\unitlength * \real{\svgscale}}%
    \fi%
  \else%
    \setlength{\unitlength}{\svgwidth}%
  \fi%
  \global\let\svgwidth\undefined%
  \global\let\svgscale\undefined%
  \makeatother%
  \begin{picture}(1,0.60265504)%
    \put(0,0){\includegraphics[width=\unitlength]{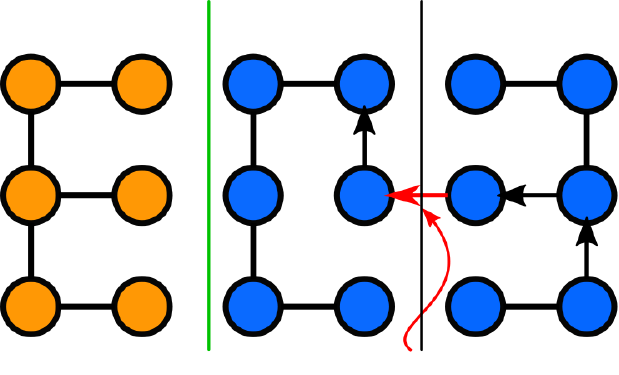}}%
    \put(0.45728807,0.00055038){\color[rgb]{1,0,0}\makebox(0,0)[lb]{\smash{root error $\rho^*(w)$: false cut}}}%
    \put(0.35149463,0.56041678){\color[rgb]{0,0.76078431,0}\makebox(0,0)[lb]{\smash{$\partial S^*$}}}%
    \put(0.32272751,0.17959165){\color[rgb]{0,0,0}\rotatebox{90}{\makebox(0,0)[lb]{\smash{$\tilde f(e) = \infty$}}}}%
    \put(0.92141868,0.09033165){\color[rgb]{0,0,0}\makebox(0,0)[lb]{\smash{$m^*$}}}%
    \put(0.57244834,0.45625572){\color[rgb]{0,0,0}\makebox(0,0)[lb]{\smash{$w$}}}%
    \put(0.69970118,0.56041678){\color[rgb]{0,0,0}\makebox(0,0)[lb]{\smash{$\partial S$}}}%
  \end{picture}%
\endgroup%

%% file: icml_watershed.bbl
\begin{thebibliography}{10}\itemsep=-1pt

\bibitem{angulo2007stochastic}
J.~Angulo and D.~Jeulin.
\newblock Stochastic watershed segmentation.
\newblock In {\em PROC. of the 8th International Symposium on Mathematical
  Morphology}, pages 265--276, 2007.

\bibitem{arbelaez_11_gpb}
P.~Arbelaez, M.~Maire, C.~Fowlkes, and J.~Malik.
\newblock Contour detection and hierarchical image segmentation.
\newblock {\em IEEE Trans. Patt. Anal. Mach. Intell.}, 33(5):898--916, 2011.

\bibitem{isbi2012challenge}
I.~Arganda-Carreras, S.~Turaga, D.~Berger, et~al.
\newblock Crowdsourcing the creation of image segmentation algorithms for
  connectomics.
\newblock {\em Front. Neuroanatomy}, 9:142, 2015.

\bibitem{bai2016deep_watershed}
M.~Bai and R.~Urtasun.
\newblock Deep watershed transform for instance segmentation.
\newblock {\em arXiv:1611.08303}, 2016.

\bibitem{beier2017multicut}
T.~Beier, C.~Pape, N.~Rahaman, T.~Prange, et~al.
\newblock Multicut brings automated neurite segmentation closer to human
  performance.
\newblock {\em Nature Methods}, 14(2):101--102, 2017.

\bibitem{cai2016pancreas}
J.~Cai, L.~Lu, Z.~Zhang, F.~Xing, L.~Yang, and Q.~Yin.
\newblock Pancreas segmentation in {MRI} using graph-based decision fusion on
  convolutional neural networks.
\newblock In {\em Proc. MICCAI}, 2016.

\bibitem{ciresan_12_deep-em-segmentation}
D.~C. Ciresan, A.~Giusti, L.~M. Gambardella, and J.~Schmidhuber.
\newblock Deep neural networks segment neuronal membranes in electron
  microscopy images.
\newblock {\em Proc. NIPS'12}, 2012.

\bibitem{powerws}
C.~Couprie, L.~Grady, L.~Najman, and H.~Talbot.
\newblock Power watershed: A unifying graph-based optimization framework.
\newblock {\em IEEE Trans. Patt. Anal. Mach. Intell.}, 33(7), 2011.

\bibitem{watershedcuts}
J.~Cousty, G.~Bertrand, L.~Najman, and M.~Couprie.
\newblock Watershed cuts: Minimum spanning forests and the drop of water
  principle.
\newblock {\em IEEE Trans. Patt. Anal. Mach. Intell.}, 2009.

\bibitem{cremi}
C.~CREMI.
\newblock Miccai challenge on circuit reconstruction from electron microscopy
  images, 2017.

\bibitem{falcao2004image}
A.~X. Falc{\~a}o, J.~Stolfi, and R.~de~Alencar~Lotufo.
\newblock The image foresting transform: Theory, algorithms, and applications.
\newblock {\em IEEE Trans. Patt. Anal. Mach. Intell.}, 26(1):19--29, 2004.

\bibitem{fowlkes_03_learning-affinity}
C.~Fowlkes, D.~Martin, and J.~Malik.
\newblock Learning affinity functions for image segmentation: combining
  patch-based and gradient-based approaches.
\newblock In {\em Proc. CVPR}, 2003.

\bibitem{grady2006random}
L.~Grady.
\newblock Random walks for image segmentation.
\newblock {\em IEEE Trans. Patt. Anal. Mach. Intell.}, 28(11):1768--1783, 2006.

\bibitem{jain2007supervised}
V.~Jain, J.~F. Murray, F.~Roth, S.~Turaga, V.~Zhigulin, K.~L. Briggman, M.~N.
  Helmstaedter, W.~Denk, and H.~S. Seung.
\newblock Supervised learning of image restoration with convolutional networks.
\newblock {\em Proc. ICCV'07}, pages 1--8, 2007.

\bibitem{floodfill}
M.~Januszewski, J.~Maitin{-}Shepard, P.~Li, J.~Kornfeld, W.~Denk, and V.~Jain.
\newblock Flood-filling networks.
\newblock {\em arXiv:1611.00421}, 2016.

\bibitem{knowles2016rhoananet}
S.~Knowles-Barley, V.~Kaynig, T.~R. Jones, A.~Wilson, J.~Morgan, D.~Lee,
  D.~Berger, N.~Kasthuri, J.~W. Lichtman, and H.~Pfister.
\newblock {RhoanaNet} pipeline: Dense automatic neural annotation.
\newblock {\em arXiv:1611.06973}, 2016.

\bibitem{kokkinos2015pushing}
I.~Kokkinos.
\newblock Pushing the boundaries of boundary detection using deep learning.
\newblock {\em arXiv:1511.07386}, 2015.

\bibitem{lillicrap2015continuous}
T.~P. Lillicrap, J.~J. Hunt, A.~Pritzel, N.~Heess, T.~Erez, Y.~Tassa,
  D.~Silver, and D.~Wierstra.
\newblock Continuous control with deep reinforcement learning.
\newblock {\em arXiv:1509.02971}, 2015.

\bibitem{VOI}
M.~Meila.
\newblock Comparing clusterings: an axiomatic view.
\newblock In {\em Proc. ICML'05}, pages 577--584, 2005.

\bibitem{meirovitch2016multi}
Y.~Meirovitch, A.~Matveev, H.~Saribekyan, D.~Budden, D.~Rolnick, G.~Odor,
  S.~K.-B. T.~R. Jones, H.~Pfister, J.~W. Lichtman, and N.~Shavit.
\newblock A multi-pass approach to large-scale connectomics.
\newblock {\em arXiv preprint:1612.02120}, 2016.

\bibitem{meyer1994minimum}
F.~Meyer.
\newblock Minimum spanning forests for morphological segmentation.
\newblock In {\em Mathematical morphology and its applications to image
  processing}, pages 77--84. 1994.

\bibitem{meyer2014watersheds}
F.~Meyer.
\newblock Watersheds on weighted graphs.
\newblock {\em Pattern Recognition Letters}, 47:72 -- 79, 2014.

\bibitem{A3C}
V.~Mnih, A.~P. Badia, M.~Mirza, A.~Graves, T.~P. Lillicrap, et~al.
\newblock Asynchronous methods for deep reinforcement learning.
\newblock In {\em Proc. ICML'16}, 2016.

\bibitem{GALA}
J.~Nunez-Iglesias, R.~Kennedy, T.~Parag, J.~Shi, and D.~Chklovskii.
\newblock Machine learning of hierarchical clustering to segment {2D} and {3D}
  images.
\newblock {\em PLoS one}, 8:e71715, 2013.

\bibitem{quan2016fusionnet}
T.~M. Quan, D.~G. Hilderbrand, and W.-K. Jeong.
\newblock {FusionNet:} a deep fully residual convolutional neural network for
  image segmentation in connectomics.
\newblock {\em arXiv:1612.05360}, 2016.

\bibitem{roerdink2000watershed}
J.~B. Roerdink and A.~Meijster.
\newblock The watershed transform: Definitions, algorithms and parallelization
  strategies.
\newblock {\em Fundamenta informaticae}, 41(1, 2):187--228, 2000.

\bibitem{ronneberger_15_u-net}
O.~Ronneberger, P.~Fischer, and T.~Brox.
\newblock {U-Net:} convolutional networks for biomedical image segmentation.
\newblock {\em Proc. MICCAI'15}, pages 234--241, 2015.

\bibitem{silver2014deterministic}
D.~Silver, G.~Lever, N.~Heess, T.~Degris, et~al.
\newblock Deterministic policy gradient algorithms.
\newblock In {\em Proc. ICML'14}, 2014.

\bibitem{sutton1999policy}
R.~S. Sutton, D.~A. McAllester, S.~P. Singh, Y.~Mansour, et~al.
\newblock Policy gradient methods for reinforcement learning with function
  approximation.
\newblock In {\em Proc. NIPS'99}, 1999.

\bibitem{MALIS}
S.~C. Turaga, K.~L. Briggman, M.~Helmstaedter, W.~Denk, and H.~S. Seung.
\newblock Maximin affinity learning of image segmentation.
\newblock {\em arXiv:0911.5372}, 2009.

\bibitem{turaga2010convolutional}
S.~C. Turaga, J.~F. Murray, V.~Jain, F.~Roth, M.~Helmstaedter, K.~Briggman,
  W.~Denk, and H.~S. Seung.
\newblock Convolutional networks can learn to generate affinity graphs for
  image segmentation.
\newblock {\em Neural Computation}, 22(2):511--538, 2010.

\bibitem{viscous2005}
C.~Vachier and F.~Meyer.
\newblock The viscous watershed transform.
\newblock {\em J. Math. Imaging and Vision}, 22(2):251--267, 2005.

\bibitem{xie2015holistically}
S.~Xie and Z.~Tu.
\newblock Holistically-nested edge detection.
\newblock In {\em Proc. ICCV'15}, pages 1395--1403, 2015.

\bibitem{DBLP:journals/corr/YuK15}
F.~Yu and V.~Koltun.
\newblock Multi-scale context aggregation by dilated convolutions.
\newblock {\em arXiv:1511.07122}, 2015.

\bibitem{zlateski2015image}
A.~Zlateski and H.~S. Seung.
\newblock Image segmentation by size-dependent single linkage clustering of a
  watershed basin graph.
\newblock {\em arXiv:1505.00249}, 2015.

\end{thebibliography}
